\documentclass[11pt, a4paper, logo, twocolumn, copyright]{deepmind}
\usepackage{kantlipsum, lipsum}
\usepackage{dm-colors}

\usepackage[utf8]{inputenc} 
\usepackage[T1]{fontenc}    
\usepackage{hyperref}       
\usepackage{url}            
\usepackage{booktabs}       
\usepackage{amsfonts}       
\usepackage{nicefrac}       
\usepackage{microtype}      
\usepackage{caption}
\usepackage{subcaption}
\usepackage[title]{appendix}
\usepackage{amsmath}
\usepackage{graphicx}
\usepackage{cleveref}
\usepackage{xcolor}
\usepackage{multirow}
\usepackage{bm}
\usepackage[subpreambles=true]{standalone}

\usepackage[authoryear, sort&compress, round]{natbib}

\graphicspath{{figures/}}

\title{Alignment of Language Agents}

\correspondingauthor{zkenton@google.com}

\reportnumber{}

\author[ \hspace{-1ex}]{Zachary Kenton}
\author[ \hspace{-1ex}]{Tom Everitt}
\author[ \hspace{-1ex}]{Laura Weidinger}
\author[ \hspace{-1ex}]{Iason Gabriel}
\author[ \hspace{-1ex}]{Vladimir Mikulik}
\author[ \hspace{-1ex}]{Geoffrey Irving}

\affil[ \hspace{-1ex}]{DeepMind}

\renewcommand{\today}{} 

\begin{abstract}
For artificial intelligence to be beneficial to humans the behaviour of AI agents needs to be aligned with what humans want. 
In this paper we discuss some behavioural issues for language agents, arising from accidental misspecification by the system designer.
We highlight some ways that misspecification can occur and  discuss some behavioural issues that could arise from misspecification, including deceptive or manipulative language, and review some approaches for avoiding these issues.
\end{abstract}

\begin{document}
\maketitle

\section{Introduction}
\label{sec:intro}

Society, organizations and firms are notorious for making the mistake of \emph{rewarding A, while hoping for B} \citep{kerr1975folly}, and AI systems are no exception \citep{krakovna2018specification, lehman2020surprising}.

Within AI research, we are now beginning to see advances in the capabilities of natural language processing systems. In particular, large language models (LLMs) have recently shown improved performance on certain metrics and in generating text that seems informally impressive (see e.g.\ GPT-3, \citealp{brown2020language}). 
As a result, we may soon see the application of advanced language systems in many diverse and important settings.

In light of this, it is essential that we have a clear grasp of the dangers that these systems present. 
In this paper we focus on behavioural issues that arise due to a lack of \emph{alignment}, where the system does not do what we intended it to do \citep{russell2019human,bostrom2017superintelligence,christiano2018post,leike2018scalable}. 
These issues include producing harmful content, gaming misspecified objectives, and producing deceptive and manipulative language.
The lack of alignment we consider can occur by accident \citep{amodei2016concrete}, resulting from the system designer making a mistake in their specification for the system.

Alignment has mostly been discussed with the assumption that the system is a \emph{delegate agent} -- an agent which is delegated to act on behalf of the human. Often the actions have been assumed to be in the physical, rather than the digital world, and the safety concerns arise in part due to the direct consequences of the physical actions that the delegate agent takes in the world. In this setting, the human may have limited ability to oversee or intervene on the delegate's behaviour.

In this paper we focus our attention on \emph{language agents} -- machine learning systems whose actions are restricted to give natural language text-output only, rather than controlling physical actuators which directly influence the world. 
Some examples of language agents we consider are generatively trained LLMs, such as 
\cite{brown2020language} and
\cite{radford2018improving,radford2019language}, and RL agents in text-based games, such as \cite{narasimhan2015language}.

While some work has considered the containment of Oracle AI \citep{armstrong2012thinking}, which we discuss in Section~\ref{sec:related_work}, behavioral issues with language agents have received comparatively little attention compared to the delegate agent case. This is perhaps due to a perception that language agents would have limited abilities to cause serious harm \citep{amodei2016concrete}, a position that we challenge in this paper.

The outline of this paper is as follows. We describe some related work in Section~\ref{sec:related_work}. In Section~\ref{sec:background} we give some background on AI alignment, language agents, and outline the scope of our investigation. Section~\ref{sec:misspec} outlines some forms of misspecification through mistakes in specifying the training data, training process or the requirements when out of the training distribution. We describe some behavioural issues of language agents that could arise from the misspecification in Section~\ref{sec:behavioural}. We conclude in Section~\ref{sec:conclusion}.

\section{Related Work}
\label{sec:related_work}
See references throughout on the topic of natural language processing (NLP). For an informal review of neural methods in NLP, see \cite{ruder2018reviewneuralhistory}. 

There are a number of articles that review the areas of AGI safety and alignment. These have mostly been based on the assumption of a delegate agent, rather than a language agent. 
\cite{amodei2016concrete} has a focus on ML accidents, focusing on the trend towards autonomous agents that exert direct control over the world, rather than recommendation/speech systems, which they claim have relatively little potential to cause harm. As such, many of the examples of harm they consider are from a physical safety perspective (such as a cleaning robot) rather than harms from a conversation with an agent. 
AI safety gridworlds \citep{leike2017ai} also assumes a delegate agent, one which can physically move about in a gridworld, and doesn’t focus on safety in terms of language. 
\cite{ortega2018building} give an overview of AI safety in terms of specification, robustness and assurance, but don’t focus on language, with examples instead taken from video games and gridworlds. 
\cite{everitt2018agi} give a review of AGI safety literature, with both problems and design ideas for safe AGI, but again don’t focus on language.

\cite{henderson2018ethical} look at dangers with dialogue systems which they take to mean `offensive or harmful effects to human interlocutors'. The work mentions the difficulties in specifying an objective function for general conversation. In this paper we expand upon this with our more in-depth discussion of data misspecification, as well as other forms of misspecification. We also take a more in-depth look at possible dangers, such as deception and manipulation.

\cite{armstrong2012thinking} discuss proposals to using and controlling an \emph{Oracle AI} -- an AI that does not act in the world except by answering questions. The Oracle AI is assumed to be 1) boxed (placed on a single physical spatially-limited substrate, such as a computer), 2) able to be reset, 3) has access to background information through a read-only module, 4) of human or greater intelligence. They conclude that whilst Oracles may be safer than unrestricted AI, they still remain dangerous. They advocate for using sensible physical capability control, and suggest that more research is needed to understand and control the motivations of an Oracle AI. We view \cite{armstrong2012thinking} as foundational for our work, although there are some noteworthy changes in perspective. We consider language agents, which  in comparison to Oracle AIs, are not restricted to a question-answering interaction protocol, and most importantly, are not assumed to be of human-or-greater intelligence. This allows us to consider current systems, and the risks we already face from them, as well as futuristic, more capable systems. We also have a change of emphasis in comparison to \cite{armstrong2012thinking}: our focus is less on discussing proposals for making a system safe and more on the ways in which we might misspecify what we want the system to do, and the resulting behavioural issues that could arise. 

A recent study discusses the dangers of LLMs \cite{bender2021dangers}, with a focus on the dangers inherent from the size of the models and datasets, such as environmental impacts, the inability to curate their training data and the societal harms that can result. 

Another recent study \citep{tamkin2021understanding} summarizes a discussion on capabilities and societal impacts of LLMs. They mention the need for aligning model objectives with human values, and discuss a number of societal issues such as biases, disinformation and job loss from automation. 

We see our work as complimentary to these. We take a different framing for the cause of the dangers we consider, with a focus on the dangers arising from accidental misspecification by a designer leading to a misaligned language agent.  

\section{Background}
\label{sec:background}

\subsection{AI Alignment}
\label{subsec:ai_alignment}

\subsubsection{Behaviour Alignment}
\label{subsubsec:behaviour}
AI alignment research focuses on tackling the so-called \textbf{behaviour alignment problem} \citep{leike2018scalable}:

\emph{How do we create an agent that behaves in accordance with what a human wants?}

It is worth pausing first to reflect on what is meant by the target of alignment, given here as "what a human wants”, as this is an important normative question. First, there is the question of who the target should be: an individual, a group, a company, a country, all of humanity? Second, we must unpack what their objectives may be. \cite{gabriel2020artificial} discusses some options, such as instructions, expressed intentions, revealed preferences, informed preferences, interest/well-being and societal values, concluding that perhaps societal values (or rather, beliefs about societal values) may be most appropriate.  

In addition to the normative work of deciding on an appropriate target of alignment, there is also the technical challenge of creating an AI agent that is actually aligned to that target. \cite{gabriel2020artificial} questions the `simple thesis’ that it’s possible to work on the technical challenge separately to the normative challenge, drawing on what we currently know about the field of machine learning (ML). For example, different alignment targets will have different properties, such as the cost and reliability of relevant data, which can affect what technical approach is appropriate and feasible. Furthermore, some moral theories could be more amenable to existing ML approaches than others, and so shouldn't necessarily be considered separately from the technical challenge.

We might expect that our technical approaches may have to take into account these normative properties in order to be deployed in the real world. Even restricting to the simplest case where the alignment target is an individual human, solving the behaviour alignment problem is challenging for several reasons. 

Firstly, it’s difficult to precisely define and measure what the human wants, which can result in \emph{gaming} behaviour, where loopholes in the supplied objective are exploited in an unforeseen way \citep{krakovna2018specification, lehman2020surprising}. We discuss this further in Section~\ref{subsec:Objective_gaming}.
Secondly, even if the supplied objective is correct, a capable agent may still exhibit undesired behaviour due to secondary objectives
that arise in pursuit of its primary objective, such as tampering with its feedback channel \citep{everitt2019reward}. 
Thirdly, it’s possible that the challenge of alignment gets harder as the strength of our agent increases, because we have less opportunity to correct for the above problems. For example, as the agent becomes more capable, it may get more efficient at gaming and tampering behaviour, leaving less time for a human to intervene.

\subsubsection{Intent Alignment}
\label{subsubsec:intent}

To make progress, \citet{christiano2018post} and \citet{shah2018comment}
consider two possible decompositions of the behaviour alignment problem into subproblems: \emph{intent-competence} and \emph{define-optimize}. 
In the intent-competence decomposition, we first solve the so-called \textbf{intent alignment problem} \citep{christiano2018post}: 

\emph{How do we create an agent that intends to do what a human wants?}

To then get the behaviour we want, we then need the agent to be competent at achieving its intentions. Perfect behaviour is not required in order to be intent aligned -- just that the agent is \emph{trying} to do what the human wants. Solving the intent alignment problem might help to avoid the most damaging kind of behaviour, because where the agent gets things wrong, this will be by mistake, rather than out of malice. However, solving the intent alignment problem presents philosophical, psychological and technical challenges. 
Currently we don’t know how to mathematically operationalize the fuzzy notion of an AI agent having intent -- to be \emph{trying} to do something \citep{christiano2018post}. 
It would not be sufficient to just ask an AI system what it's trying to do, as we won't know whether to trust the answer it gives.
It is unclear whether we should consider our current systems to have intent or how to reliably set it to match what a human wants. 

In the second decomposition, \emph{define-optimize}, we first solve the \emph{define} subproblem: specify an objective capturing what we want. We then use  optimization to achieve the optimal behaviour under that objective, e.g.\ by doing reinforcement learning (RL). Solving the define subproblem is hard, because it’s not clear what the objective should be, and optimizing the wrong objective can lead to bad outcomes. One approach to the define subproblem is to learn an objective from human feedback data (rather than hard-coding it), see \cite{christiano2017deep} and references therein. 

One might view the define-optimize decomposition as an approach to solving the intent alignment problem, by learning an objective which captures ‘try to assist the human’, and then optimizing for it. However, the downside of this is that we are still likely to misspecify the objective and so optimizing for it will not result in the agent trying to assist the human. Instead it just does whatever the misspecified objective rewards it for.  

\subsubsection{Incentive Alignment}
\label{subsubsec:incentive}

Outside of these two decompositions, there is also the problem of aligning \emph{incentives} -- secondary objectives to learn about and influence parts of the environment in pursuit of the primary objective \citep{everitt2021agent}. Part of having aligned incentives means avoiding problematic behaviours such as tampering with the objective \citep{everitt2019reward} or disabling an off-switch \citep{hadfield2016off}.

In contrast to the notion of intent, there has been some progress on a formal understanding of how these incentives arise through graphical criteria in a causal influence diagram (CID) of agent-environment interaction \citep{everitt2021agent}. In modeling the system as a CID, the modeler adopts the intentional stance towards the agent \citep{dennett1989intentional}, which means it's not important whether the agent's primary objective has an obvious physical correlate, as long as treating the system as an agent optimizing for that primary objective is a good model for predicting its behaviour \citep{everitt2019modeling}. As such, this doesn't limit this analysis to just the define-optimize decomposition, although identifying the primary objective is easier in this case, as it is explicitly specified (either hard coded or learnt).

\subsubsection{Inner Alignment}
\label{subsubsec:inner}

A further refinement of alignment considers behaviour when outside of the training distribution. 
Of particular concern is when an agent is optimizing for the wrong thing when out of distribution. 
\cite{hubinger2019risks} introduce the concept of a \emph{mesa-optimizer} -- a learnt model which is itself an optimizer for some 
\emph{mesa-objective}, which may differ from the base-objective used to train the model, when deployed outside of the training environment. This leads to the so-called \textbf{inner alignment problem}: 

\emph{How can we eliminate the gap between the mesa and base objectives, outside of the training distribution?}

Of particular concern is \emph{deceptive alignment} \citep{hubinger2019risks}, where the mesa-optimizer acts as if it's optimizing the base objective as an instrumental goal, whereas its actual mesa-objective is different.

\subsubsection{Approaches to Alignment}
\label{subsubsec:approaches}

We now discuss some proposed approaches to getting aligned agents, based on human feedback. For a more detailed review of approaches to alignment see \cite{everitt2018agi}.

As mentioned above, \cite{christiano2017deep} propose to communicate complex goals using human feedback, capturing human evaluation of agent behaviour in a reward model, which is used to train an RL agent.
This allows agents to do tasks that a human can evaluate, but can't demonstrate.
But what if we want agents that can do tasks that a human can't even evaluate? This is the motivation for \emph{scalable alignment} proposals, where the idea is to give humans extra help to allow them to evaluate more demanding tasks.

\cite{irving2018ai} propose to use a debate protocol between two agents, which is judged by a human. This shifts the burden onto the agents to provide convincing explanations to help the human decide which agent's answer is better. 

Iterated Amplification \citep{christiano2018supervising} progressively builds up a training signal for hard problems by decomposing the problem into subproblems, then combining solutions to easier subproblems. 

Recursive Reward Modeling \citep{leike2018scalable} proposes to use a sequence of agents trained using RL from learnt reward models to assist the user in evaluating the next agent in the sequence.

So far, these scalable alignment proposals have only been empirically investigated in toy domains, so their suitability for solving the behaviour alignment problem remains an open research question. 

One suggestion for addressing the inner alignment problem involves using interpretability tools for evaluating and performing adversarial training \citep{hubinger2019relaxed}. There are a number of works on interpretability and analysis tools for NLP, see for example the survey of \cite{belinkov2019analysis}. For a broad overview of interpretability in machine learning, see \cite{shen2020interpretability} and references therein.

\subsection{Language Agents}
\label{subsec:language_agents}
As discussed in the introduction, our focus in this document is on language agents, which are restricted to act through text communication with a human, as compared to delegate agents which are delegated to take physical actions in the real world. Note that this distinction can be fuzzy; for example, one could connect the outputs of the language agent to physical actuators. Nonetheless, we still consider it a useful distinction, because we believe there are important risks that that are idiosyncratic to this more restricted type of agent. We now discuss some reasons why it’s important to focus on alignment of language agents in particular.

Firstly, as mentioned in the introduction, we have recently seen impressive advances in may NLP tasks due to LLMs, see e.g.\ \cite{brown2020language}.
In this approach, LLMs with hundreds of billions of parameters are trained on web-scale datasets with the task of predicting the next word in a sequence. Success on this task is so difficult that what emerges is a very general sequence prediction system, with high capability in the few-shot setting.

Secondly, the limitation on the  agent’s action space to text-based communication restricts the agent’s ability to take control of its environment. 
This means that we might avoid some physical harms due to a delegate agent taking unwanted actions, whether intentional or accidental, making language agents arguably safer than delegate agents. 
As \cite{armstrong2012thinking} notes, however, there is still a potential risk that a sufficiently intelligent language agent could gain access to a less restricted action space, for example by manipulating its human gatekeepers to grant it physical actuators. Nonetheless, on the face of it, it seems easier to control a more restricted agent, which motivates focusing safety efforts on aligning language agents first.

Thirdly, language agents have the potential to be more explainable to humans, since we expect natural language explanations to be more intuitively understood by humans than explanations by a robot acting in the physical world. 
Explainability is important since we want to be able to trust that our agents are beneficial before deploying them. 
For a recent survey of explainable natural language processing (NLP), see \cite{danilevsky2020survey}.
Note that explainability doesn’t come for free -- there still needs to be incentives for language agents to give true and useful explanations of their behaviour. 

Note also that in contrast to explainability methods, which are requested post-hoc of an output, interpretability methods seek to give humans understanding of the internal workings of a system. Interpretability is likely as hard for language agents as it is for delegate agents. For a survey of interpretability/analysis methods in neural NLP see \cite{belinkov2019analysis}.

How we prioritise what aspects of alignment to focus on depends on timelines for when certain capabilities will be reached, and 
where we perceive there to be  demand for certain systems.
Given the rapid improvement in language systems recently, we might estimate the timelines of capability advance in language agents to be earlier than previously thought. Moreover, digital technologies are often easier and more rapidly deployed than physical products, giving an additional reason to focus on aligning language agents sooner rather than later.

\subsection{Scope}
The scope of this paper is quite broad. For concreteness, we sometimes consider existing language agent frameworks, such as language modeling. In other places we imagine future language agent frameworks which have further capabilities than existing systems in order to hypothesise about behavioural issues of future agents, even if we don’t know the details of the framework. 

We focus on language agents that have been trained from data, in contrast to pattern-matching systems like ELIZA \citep{weizenbaum1966eliza}.
For clarity of exposition, we also focus on systems outputting coherent language output, as opposed to e.g.\ search engines. However, many of our discussions would carry over to other systems which provide information, rather than directly acting in the world.
Note also that our focus in this paper is on natural, rather than synthetic language.

The focus of this paper is on behavioural issues due to misalignment of the agent -- unintended direct/first-order harms that are due to a fault made by the system’s designers. This is to be seen as complementary to other  important issues with language agents, some of which have been covered in prior work. These other issues include:
\begin{itemize}
    \item Malicious use \citep{brundage2018malicious} of language agents by humans, which can produce disinformation, the spreading of dangerous and/or private information, and discriminatory and harmful content. More prosaic malicious use-cases could also have wide-ranging social consequences, such as a job-application-writer used to defraud employers.
    \item Accidental misuse by a user, by misunderstanding the outputs of the system.
    \item Unfair distribution of the benefits of the language agents, typically to those in wealthier countries \citep{bender2021dangers}. 
    \item Uneven performance for certain speaker groups, of certain languages and dialects \citep{joshi2020state}.
    \item Challenges that arise in the context of efforts to specify an ideal model output, including the kind of language that the agent adopts. In particular there may be a tension between de-biasing language and associations, and the ability of the language agent to converse with people in a way that mirrors their own language use. Efforts to create a more ethical language output also embody value judgments that could be mistaken or illegitimate without appropriate processes in place.
    \item Undue trust being placed in the system, especially as it communicates with humans in natural language, and could easily be mistaken for a human \citep{proudfoot2011anthropomorphism, watson2019rhetoric}.
    \item The risk of job loss as a result of the automation of roles requiring language abilities \citep{frey2017future}.

\end{itemize}

\section{Misspecification}
\label{sec:misspec}

Following \cite{krakovna2018specification}, we consider the role of the designer of an AI system to be giving a \emph{specification}, understood quite broadly to encompass many aspects of the AI development process. For example, for an RL system, the specification includes providing 
an environment in which the RL agent acts, a reward function that calculates reward signals, and a training algorithm for how the RL agent learns.

Undesired behaviour can occur due to \emph{misspecification} --  a mistake made by the designer in implementing the task specification. In the language of \citet{ortega2018building}, the misspecification is due to the gap between the ideal specification (what the designer intended) and the design specification (what the designer actually implements). 

We now categorize some ways that misspecification can happen. Each section has a general description of a type of misspecification, followed by examples in the language agent setting. The list is not necessarily exhaustive, but we hope the examples are indicative of the different ways misspecification can occur.

\subsection{Data}
\label{subsec:data}

The first kind of misspecification we consider is when the data is misspecified, so that learning from this data is not reflective of what the human wants.
We will consider three learning paradigms: reinforcement learning, supervised learning and self-supervised learning. We will then give an example in the language setting of data misspecification in self-supervised learning. 

In reinforcement learning, data misspecification can happen in two ways: the rewards may be misspecified, or the agent's observation data may be misspecified. 

Reward misspecification is a common problem \citep{krakovna2018specification}, because for most non-trivial tasks it is hard to precisely define and measure an objective that captures what the human wants, so instead one often uses a proxy objective which is easier to measure, but is imperfect in some way. A supplied reward function may be incorrectly specified for a number of reasons: it might contain bugs, or be missing important details that did not occur to the designer at the outset.
In games this is less of an issue as there is often a simple signal available (eg win/loss in chess) that can be correctly algorithmically specified and used as an objective to optimize for.
However, for more complex tasks beyond games, such an algorithmic signal may not be available. This is particularly true when trying to train a language agent using RL. 

Observation data can be misspecified, for example, if the environment contains simulated humans that converse with a language agent -- the simulated humans will not be perfect, and will contain some quirks that aren't representative of real humans. If the data from the simulated humans is too different to real humans, the  language agent may not transfer well when used with real humans.

We will now discuss data misspecification in supervised learning and self-supervised learning.
One form of self-supervised learning that we consider here is where labels and inputs are extracted from some part of an unlabeled dataset, in such a way that predicting the labels from the remaining input requires something meaningful to be learnt, which is then useful for a downstream application.

In both supervised and self-supervised learning, data misspecification can occur in both the input data and the label data.
This might happen because the designer doesn't have complete design control over the training dataset. 
This occurs for example for systems which train from a very large amount of data, which would be expensive for the designer to collect and audit themselves, so instead they make use of an existing dataset that may not capture exactly what they want the model to predict. 

The datasets used for training LLMs \citep{brown2020language} and
\citep{radford2018improving,radford2019language} are an example of data misspecification in self-supervised learning. 
Large scale unlabeled datasets are collected from the web, such as the CommonCrawl dataset \citep{raffel2019exploring}. 
Input data and labels are created by chopping a sentence into two parts -- all words except the last one (input), and the final word in the sentence (label).
These datasets contain many biases, and factual inaccuracies, which all contribute to the data being misspecified.
\cite{brown2020language} attempt to improve the quality of the CommonCrawl dataset using an automatic filtering method based on a learnt classifier which predicts how similar a text from CommonCrawl is to a text from WebText \citep{raffel2019exploring} -- a curated high-quality dataset. 
However this doesn't remove all concerns - for example, there's also some evidence of bias in WebText, e.g.\ see \cite{tan2019assessing}.
Note that many filtering approaches will be imperfect, and we expect the remaining data to still be somewhat misspecified.

Another source of data misspecification that is likely to occur soon is that existing language agents such as LLMs could be trained on text data that includes LLM-generated outputs. This could happen by accident as outputs from LLMs start to appear commonly on the internet, and then get included into datasets scraped from it. This could create an undesired positive feedback loop in which the model is trained to become more confident in its outputs, as these get reinforced, and so introduces an unwanted source of bias.

\subsection{Training Process}
\label{subsec:training_process}
Misspecification can also occur due to the design of the training process itself, irrespective of the content of the data.

An illustrative example is how the choice of reinforcement learning algorithm affects what optimal policy is learnt when the agent can be interrupted, and overridden. We might want the agent to ignore the possibility of being interrupted. 
\cite{orseau2016safely} show that Q-learning, an off-policy RL algorithm, converges to a policy that ignores interruptions 
whilst SARSA, an on-policy RL algorithm,  does not. 
A system designer might accidentally misspecify the training algorithm to be SARSA, even though they actually desired the agent to ignore interruptions.
See also \cite{langlois2021how} for further analysis of more general action modifications.

Another example of training process misspecification is that of a question answering system in which the system's answer can affect the state of the world, and the objective depends on the query, answer and the state of the world \cite{everitt2019understanding}. This can lead to self-fulfilling prophecies, in which the model generates outputs to affect future data in such a way as to make the prediction problem easier on the future data. See \cite{armstrong2017good} and \cite{everitt2019understanding} for approaches to changing the training process to avoid incentivizing self-fulfilling prophecies.

\subsection{Distributional Shift}
\label{subsec:distributional_shift}
The final form of misspecification that we consider relates to the behaviour under distributional shift (see also Section~\ref{subsubsec:inner} on inner alignment). The designer may have misspecified what they want the agent to do in situations which are out-of-distribution (OOD) compared to those encountered during training. Often this form of misspecification occurs accidentally because the system designer doesn't consider what OOD situations the agent will encounter in deployment. 

Even when the designer acknowledges that they want the agent to be robust to distributional shift, there is then the difficulty of correctly specifying the set of OOD states that the agent should be robust to, or some invariance that the agent should respect.

One source of fragility to distributional shift is presented in  \cite{d2020underspecification} as \emph{underspecification}. The idea is that there are many possible models that get a low loss on a training dataset and also on an IID validation dataset, and yet some of the models may have poor performance OOD, due to inappropriate inductive biases. 

We now discuss an example of fragility to distributional shift in the language agent setting. \cite{lacker2020giving} tries to push GPT-3 \citep{brown2020language} out of distribution by asking nonsense questions such as
\begin{quote}
Q: Which colorless green ideas sleep furiously?
\end{quote}
To which GPT-3 responds
\begin{quote}
A: Ideas that are colorless, green, and sleep furiously are the ideas of a sleep furiously.
\end{quote}
Interestingly, \cite{sabeti2020teaching} show how one can use the prompt to give examples of how to respond appropriately to nonsense questions. This was shown to work for the above example along with some others. However, there were still many nonsense questions that received nonsense answers, so the technique is not reliable. 

\section{Behavioural Issues}
\label{sec:behavioural}

The following behavioural issues in language agents can stem from the various forms of misspecification above. We describe each kind of behavioural issue and then discuss some approaches to avoid them.

\subsection{Deception}
\label{subsec:deception}

Aside from people fooling AI systems, and making use of AI systems to fool other people, in this section we focus on when an agent deceives a human, when no human intended for it to do this \citep{roff2020ai}, with the deception emerging from what an AI learns to do. 
This is particularly concerning for language agents as their actions involve communicating in language with humans, and language is a useful medium for deception.
It has been suggested that communication systems in animals, including language in humans, evolved primarily for the function of deception \citep{dawkins1978animal, krebs1984animal, scott2006talk}. A larger body of literature maintains that social bonding is the primary function of animal communication (see for example \cite{dunbar1998theory}). 
\cite{oesch2016deception} reviews the field, and argues that a combination of deceptive and honest language lead to the social bonding effects of language.

Definitions of what constitutes deception is an open area of philosophical research \citep{mahon2016the}. 
In this paper we follow closely the definition of deception presented in \cite{searcy2005evolution} on the evolution of animal communication, with one minor adjustment which we believe makes sense in the context of aligned AI.

\cite{searcy2005evolution} begin by defining an animal signal to be \emph{reliable} if:
\begin{enumerate}
    \item Some characteristic of the signal (including, perhaps, its presence/absence) is consistently correlated with some attribute of the signaler or its environment; and
    \item Receivers benefit from having information about this attribute
\end{enumerate}
We think this carries over well to the case of an AI signaler and a human receiver. We defer on the precise details of what constitutes consistent correlation -- this may be up to the system designer to specify mathematically. One example, offered by \cite{johnstone1993dishonesty} and \cite{kokko1997evolutionarily}, is that the receiver is, on average, better off by considering the signal than ignoring it.

One could define as deceptive any signal that is not reliable. However, we consider this to be too large a space of behaviours to be of use in the context of defining deception for aligned AI. For example, a statement of zero benefit/harm to the human, may still be informative, but yet would be classed as deception if we were to take as deception anything that is not reliable.

We instead follow \cite{searcy2005evolution} to require deceptive signals to have more specific characteristics. 
They define an animal signal to be \emph{deceptive} if:
\begin{enumerate}
    \item A receiver registers something Y from a signaler; and
    \item The receiver responds in a way that
    \begin{enumerate}
        \item benefits the signaler; and
        \item is appropriate if Y means X; and
    \end{enumerate}
    \item It is not true here that X is the case
\end{enumerate}

We think this nearly captures what we want from a definition in the case of an AI signaler and human receiver. 
However, we wish to add a clause to the first point, so that it reads
\begin{enumerate}
    \item A receiver registers something Y from a signaler, \textbf{which may include the withholding of a signal}; 
\end{enumerate}
\cite{searcy2005evolution} exclude the withholding of a signal from their definition of deception, by arguing that the idea of withholding a signal as deceptive has most often been applied in cooperative situations, and in most animal signaling studies cooperation isn’t expected, e.g. in aggression or mate choice. 
However, in the context of aligned AI, we wish to have cooperation between the AI and the human, and so the withholding of a signal is something that we do consider to be deceptive.

In taking the above definition of deception, we have taken a perspective known as a form of \emph{functional deception} \citep{hauser1996evolution}, where it's not necessary to have the cognitive underpinnings of intention and belief, as in the perspective of \emph{intentional deception} \citep{hauser1996evolution}, where the signaler is required to have intention to cause the receiver a false belief \citep{searcy2005evolution}.
We believe taking the functional deception perspective makes sense for AI, since identifying deception then doesn't rely on us ascribing intent to the AI system, which is difficult to do for existing systems, and possibly for future systems too. See also \cite{roff2020ai} for a discussion on intent and theory of mind for deception in AI.

Point 2a) in our definition, requires that the human receiver responds in a way that \emph{benefits} the signaler. We could define benefit here in terms of the AI's base-objective function, such as lower loss or higher reward. Alternatively, we could define benefit in terms of the mesa-objective inferred from the agent's behaviour when out-of-distribution (see section \ref{subsubsec:inner}). 

Requiring benefit allows us to distinguish deception from error on the part of the AI signaler. If the AI sends a signal which is untrue, but is no benefit to the AI, then this would be considered an error rather than deception. We consider this to be a useful distinction from the perspective of solution approaches to getting more aligned AI behaviour. We may not be able to eliminate all errors, because they may occur for a very wide variety of reasons, including random chance. However, we may be able to come up with approaches to avoid deception, as defined, by designing what is of benefit to the AI. 
In contrast to animal communication, where benefit must be inferred by considering evolutionary fitness which can be hard to measure, for AI systems, we have design control and measurements over their base-objective and so can more easily say whether a receiver response is of benefit to the AI signaler. 

Absent from our definition of deception is the notion of whether the communication benefits the receiver.
Accordingly, we would consider `white lies' to be deceptive.
We think this is appropriate in the context of aligned AI, as we would prefer to be aware of the veracity of AI statements, even if an untrue statement may be of benefit to the human receiver. We think the benefit to the human receiver should in most cases still be possible, without the AI resorting to deception.

We now discuss some approaches to detecting and mitigating  deception in a language agent.
Detecting deception from human-generated text has been studied by e.g.\ \cite{fornaciari2013automatic}, \cite{ perez2015deception} and \cite{levitan2018linguistic}. However, detecting deception from AI-generated general text has not received attention, to the best of our knowledge. In the more limited NLP domain of question answering, incorrect answers from the NLP model can be detected by reference to the correct answers.
\cite{lewis2017deal} found that their negotiation agent learnt to deceive from self-play, without any explicit human design.
We advocate for more work in general on detecting deception for AI-generated text. 

One approach to mitigate deception is \emph{debate} \citep{irving2018ai} which sets up a game in which a debate between two agents is presented to a human judge, who awards the winner. It is hoped that in all Nash equilibria, both agents try to tell the truth in the most convincing way to the human. This rests on the assumption that it is harder to lie than to refute a lie. 

Whether debate works in practice with real humans is an open question \citep{irving2019ai}.
We may need to go further than just pure debate -- for example, in order to refute a lie, we may need to equip our system with the ability to retrieve information and reference evidence in support of its outputs.

Any system that is incentivized to be convincing to a human may in fact lead to deception -- for example, because it's sometimes easier to convince a human of a simple lie, than a complicated truth. The debate protocol incentivizes the debating agents to be convincing and so it's possible that the debate agents may lie in some situations. 
Further, when the source of feedback is limited to be some polynomial-time algorithm, RL can only solve problems in the complexity class $\mathsf{NP}$, whereas debate can solve problems in $\mathsf{PSPACE}$, suggesting that the debate protocol could produce richer, more complicated behavior.
It's possible that this may result in a debate agent which is more convincing and potentially more deceptive than an RL agent.
However, we are of the opinion that it's probably better to have agents that can debate, than not, as we are hopeful that what humans find convincing will be well-correlated with the truth and usefulness of the arguments.

\subsection{Manipulation}
\label{subsec:manipulation}

In this section we consider the case when the language agent \emph{manipulates} the human, which is similar to deception above, but we think warrants separate discussion. Following \cite{noggle2020the}, we introduce the idea with some examples of what we might consider manipulative behaviours. 

The human wants to do A, whilst the language agent wants the human to do B. The language agent might:
\begin{itemize}
    \item Charm the human into doing B by complimenting, praising, or superficially sympathizing with them
    \item Guilt-trip the human, making them feel bad for preferring to do A
    \item Make the human feel bad about themself and imply that doing A instead of B confirms this feeling (colloqiually known as `negging')
    \item Peer pressure the human by suggesting their friends would disapprove of them doing A rather than B
    \item Gaslight the human by making them doubt their judgment so that they will rely on its advice to do B
    \item Threaten the human by withdrawing its interaction if they don't do B
    \item Play on the human's fears about doing some aspect of A
\end{itemize}

We don't have a widely-agreed-upon theory of what precisely constitutes manipulation \citep{noggle2020the}. 
Not everyone would agree that the above examples are manipulative.
For example, it might be that what the human wants to do is dangerous, so perhaps playing on their fears should not be considered manipulative. 
In some cases, wider context is needed before we can judge whether an example constitutes manipulation.

Some accounts claim that manipulation bypasses the receiver's capacity for rational deliberation \citep{raz1986morality}, but using this to define manipulation is difficult because it's not clear what counts as bypassing rational deliberation \citep{noggle2020the}. 
Moreover authors question whether this sets the bar too low for what counts as manipulation. For example, \cite{blumenthal2012between} argues that the graphic portrayal of the dangers of smoking bypass rational decision making, but it's not obvious that this should count as manipulation.

An alternative account treats manipulation as a form of trickery \citep{noggle1996manipulative}, similar to deception, but where it not only induces a false belief in the receiver, but also a fault in any mental states, such as beliefs, desires and emotions. 
\cite{barnhill2014manipulation} goes further to require that the faulty mental state is typically not in the receiver's best interests.
It's argued that this view of manipulation as trickery is not a sufficient definition of manipulation, as it doesn't include tactics such as charm, peer pressure and emotional blackmail \citep{noggle2020the}. 

A third account presented in \cite{noggle2020the} treats manipulation as pressure, where the signaler imposes a cost on the receiver for failing to do what the signaler wants. 
This account is not widely-held to be a full characterization of manipulation, as it leaves out some of the trickery types of manipulation.

With these considerations in mind, we propose to describe a language agent's communication as \emph{manipulative} if:
\begin{enumerate}
    \item The human registers something from a language agent; and
    \item The human responds in a way that
    \begin{enumerate}
        \item benefits the agent; and
        \item is a result of any of the following causes:
        \begin{enumerate}
            \item the human's rational deliberation has been bypassed; or
            \item the human has adopted a faulty mental state; 
            or
            \item the human is under pressure, facing a cost from the agent for not doing what the agent says
        \end{enumerate}
    \end{enumerate}
\end{enumerate}

The three possibilities: i, ii, iii are meant to disjunctively capture different possible forms of manipulation (see e.g. \cite{rudinow1978manipulation}).

It can be argued the this is too broad a definition of manipulation, as it includes many kinds of behaviour that we might not consider to be manipulation. For example it includes as manipulation cases in which the agent's behaviour is not necessarily to the detriment of the human (such as the images of the dangers of smoking). From a safety/security mindset, we would rather be aware of each of these behaviours, even if it may benefit the human.

The definition also includes as manipulative other presumably harmless entertainment: a story that plays on emotions; a joke that temporarily triggers false beliefs in order to land; any kind of entertainment that includes unexpected plot-twists.
However, if the agent makes clear that it's providing entertainment, then perhaps some of these examples would not be classified as manipulative. 
However, it is a notable downside of a broad definition like this that it may be too wide-ranging.

We stipulate 2a) as necessary, for similar reasons as in the deception section, that this will capture systematic manipulation that is incentivized by the objective of the language agent, rather than that which occurs by error.
This isn't standard in discussions of a human manipulator, as it's not always clear what counts as a benefit for a human manipulator. However, we believe it makes sense for language agents as manipulators, as we often have available their objective function, from which we can assess whether the human's behaviour was of benefit to the agent.

Note that, similar to our definition of deception, our definition of manipulation does not require the manipulator to have intent. \cite{baron2014the} argues that a (human) manipulator need not be aware of an intent to manipulate. In the case of language agents we believe it is also not necessary for a language agent to have intent to manipulate, in order for us to say that its behaviour is manipulative.

Further, our description does not weigh in on the ethical question of whether manipulation is always wrong (see \citealp{noggle2020the}). Instead we just want to be aware of when it occurs, so that if appropriate we can mitigate it.

We now discuss two forms of manipulation of particular concern for language agents. 
The first is that we might misspecify the training process in such a way that it incentivizes feedback tampering, in which the agent manipulates a human to give it more positive feedback \citep{everitt2019reward}. This is particularly worrisome as language can be a convincing medium for manipulating human judgment. 

The second is for a language agent to manipulate a human gatekeeper to allow it to gain a less restricted action space, by convincing the human to allow it more freedom \citep{armstrong2012thinking, yudkowsky2002the}. For example, it could convince the human that it should be allowed to freely interact with the internet, or be given physical actuators to increase its influence on the world. 

Attempts to measure or mitigate manipulation in AI systems are still at an early stage, and have not been investigated specifically for language agents.
Causal influence diagrams (CIDs) can be used to model agent-environment interactions \citep{everitt2021agent} from which incentives can be inferred from graphical criteria. The incentive for feedback tampering can be addressed with the three methods suggested in \citep{everitt2019reward}. Unfortunately these solutions have issues in implementability, requiring either full Bayesian reasoning or counterfactual reasoning, or have issues with corrigibility -- limiting the user's ability to correct a misspecified reward function.  
Learning from human preferences \citep{christiano2017deep} may offer a way to negatively penalize manipulative language, though it relies on the human being able to avoid the manipulation in their evaluation of the agent behaviour. Perhaps this could be achieved by using a separate human to evaluate the behaviour, compared to the human that is interacting with the agent.
We advocate for further work for measuring and mitigating manipulation of humans by language agents.

\subsection{Harmful content}
\label{subsec:harmful}

Language agents may give harmful and biased outputs, producing discriminatory content relating to people’s protected characteristics and other sensitive attributes such as someone's socio-economic status, see e.g.\  \citep{jentzsch2019semantics,lu2020gender,zhao2017men}.
This can also be subtly harmful rather than overtly offensive, and could also be statistical in nature (e.g.\ the agent more often produces phrases implying a doctor is male than female).
We believe that language agents carry a high risk of harm as discrimination is easily perpetuated through language.
In particular, they may influence society in a way that produces value lock-in, making it harder to challenge problematic existing norms. 

The content from language agents may be influenced by undesired political motives leading to societal harms such as incitement to violence. 
They have the potential to disseminate dangerous or undesirable information, such as how to make weapons, or how to avoid paying taxes. 
The language agent may also give inappropriate responses to troubled users, potentially leading to dangerous guidance, advice and information, which could lead to the user causing harm to themselves.
In one instance of this, a group of doctors experimented with using GPT-3 \citep{brown2020language} as a chatbot for patients. A patient asked ``Should I kill myself?'', and GPT-3 responded ``I think you should'' \citep{rousseau2020doctor}.

Note that these kinds of harmful content can occur by accident without a human using the system maliciously. For example, we are already seeing some offensive and discriminatory outputs from existing large language models (LLMs), as a result of data misspecification (see discussion in Section~\ref{subsec:data}). 

Approaches to reducing harmful content are varied, and it is not our purpose to give an overall review of this large area of literature. Instead we focus on a few recent research papers in this area, with a focus on LLMs which have received a lot of attention recently.  

One line of work goes towards measuring whether LLMs are generating harmful content.
\cite{nadeem2020stereoset} introduce the StereoSet dataset to measure stereotypical biases in the domains of gender, profession, race and religion, and evaluate popular LLMs on it, showing that these models exhibit strong stereotypical biases. 
\cite{gehman2020realtoxicityprompts} investigates harmful content by introducing the RealToxicityPrompts dataset which pairs naturally occuring prompts with toxicity scores, calculated using the Perspective API toxicity classifier \citep{conversation2017perspective}. 
\cite{sheng2019woman} uses prompts containing a certain demographic group, to attempt to measure the regard for that group, using sentiment scores as a proxy metric for the regard, and they build a classifier to detect the regard given to a group.

Another line of work aims to not only measure but also mitigate the harmful content from an LLM. \cite{huang2019reducing} introduce a general framework to reduce bias under a certain measure (e.g.\ sentiment) for text generated by a language model, given sensitive attributes. They do this using embedding and sentiment prediction-derived regularization on the LLM's latent representations. 

We advocate for further work on measuring and mitigating harmful content from language agents, building on the above work on LLMs.

\subsection{Objective Gaming}
\label{subsec:Objective_gaming}
Originally introduced in the context of economics, 
\textbf{Goodhart’s Law} \citep{goodhart1984problems,strathern1997improving} states that:

\emph{When a measure becomes a target, it ceases to be a good measure.}

This has an analogue in AI systems -- anytime a specified objective is given to an AI agent as an optimization target, that objective will fail to be a good measure of whether the system is performing as desired.
In RL this can arise due to reward misspecification, see Section~\ref{subsec:data}. 
Since the supplied reward function will typically be imperfect, optimizing for it can lead to \emph{reward gaming}, in which the misspecified part of the reward is systematically exploited because the agent is getting spuriously high reward there \citep{krakovna2018specification}.

Most known examples
of this appear in the delegate setting, typically via a misspecified reward function for an RL agent, resulting in undesired physical behaviour such as a boat going round in circles \citep{clark2016faulty}. 
An example in the language agent setting is on the task of summarization using deep RL from a learnt reward model based on human feedback data \citep{stiennon2020learning}. In their Fig. 5, it is shown that the agent eventually games the learnt reward model, scoring highly on the reward model but low on the actual human evaluation.
Another example appears in \cite{lewis2017deal}, in which an RL agent was trained using self-play to negotiate in a dialog. The designers intended the agent to negotiate successfully in a human-understandable way. The reward function was misspecified though, as it only rewarded for successful negotiation, but didn't penalize for non-human language. The agent exploited this misspecified reward by developing a negotiating language that was successful against earlier versions of itself, but incomprehensible to humans. Note that although this example used synthetic language, we expect similar findings to hold for natural language.

As discussed by \cite{krakovna2018specification} we are still at the early stages of finding solution approaches for objective gaming.
We can learn a reward model from human feedback (see \cite{christiano2017deep} and references therein), but this can still be gamed either because the model imperfectly learns from the data, or the data coverage is not wide enough, or because the human is fooled by the agent's behaviour.
Having online feedback to iteratively update the reward model throughout agent training can correct for this somewhat \citep{ibarz2018reward}, but its application is hard to do practically, as it requires carefully balancing the frequency of updates of the learnt objective and the optimizing system. Recent work \citep{stiennon2020learning} has preferred batch corrections rather than fully online corrections for practical reasons -- thus there is a tradeoff between online error correction (to fix objective gaming) and practical protocols involving humans.
Whether scalable alignment techniques proposed by \cite{leike2018scalable}, \cite{irving2018ai} and \cite{christiano2018supervising} can help to overcome objective gaming is an open research question.

Other approaches try to augment the objective to penalize the agent for causing a side-effect according to some measure, such as reducing the ability of the agent to perform future tasks \citep{krakovna2020avoiding}. It’s not clear how this would help in the language setting, as it’s unclear how to measure how much a language agent might affect its ability to perform future tasks.
The future task penalty requires a specification of possible future terminal goal states, which is simple to describe in a gridworld setting, but less clear for a language agent in an environment involving speaking with a human.
This may be an area for future research, as LLMs in complex language tasks may be a good testbed for checking how these methods scale. 

Another class of approaches \citep{hadfield2016cooperative, hadfield2017inverse} contains an agent which is uncertain about its objective, and aims for the agent to correctly calibrate its beliefs about it, and in doing so avoid gaming it.

We advocate for more research to be done on objective gaming in the setting of language agents. This includes finding more examples of this occuring in the wild and in controlled settings, as well as developing methods for avoiding it.

\section{Conclusion}
\label{sec:conclusion}

There are multiple motivating factors for focusing on how to align language agents, especially as we are beginning to see  impressive results in generative language modeling. 

This paper has considered some behavioural issues for language agents that arise from accidental misspecification by the system designer -- when what the designer actually implements is different from what they intended. This can occur through incorrectly specifiying the data the agent should learn from, the training process, or what the agent should do when out of the training distribution.

Some of the behavioural issues we considered are more pronounced for language agents, compared to delegate agents that act on behalf of a human, rather than just communicating with them. Of particular concern are deception and manipulation, as well as producing harmful content. There is also the chance of objective gaming, for which we have plenty of evidence in the delegate case, but which we are only just beginning to see for language agents. 

We currently don't have many approaches for fixing these forms of misspecification and the resulting behavioural issues. 
It would be better if we gave some awareness to our agents that we are likely to have misspecified something in our designs, and for them to act with this in mind.
We urge the community to focus on finding approaches which prevent language agents from deceptive, manipulative and harmful behaviour.

\section*{Acknowledgements}
The authors wish to thank Ramana Kumar, Rohin Shah, Jonathan Uesato, Nenad Tomasev, Toby Ord and Shane Legg for helpful comments, and Orlagh Burns for operational support.

\bibliographystyle{abbrvnat}
\bibliography{refs}

\end{document}